\documentclass[runningheads]{llncs}
\usepackage[T1]{fontenc}
\usepackage{graphicx}
\usepackage[colorlinks=true, linkcolor=blue, urlcolor=blue, citecolor=blue]{hyperref}
\usepackage{xcolor}
\definecolor{customgreen}{HTML}{00A000} % 定义名为 customgreen 的绿色
\usepackage{tcolorbox}
\usepackage{booktabs}
\usepackage{url} 
\usepackage{tabularx}
\usepackage{authblk}
\usepackage{pifont} 
\usepackage{afterpage}

\usepackage{multirow}
\usepackage{caption}
\usepackage{geometry}
\usepackage{adjustbox}
\usepackage{amsmath}
\usepackage{amssymb}
\usepackage{amsfonts}
\usepackage{enumitem}

\begin{document}
\pagestyle{plain}

\title{PoseLLM: Enhancing Language-Guided Human Pose Estimation with MLP Alignment}
%
% \titlerunning{}
% If the paper title is too long for the running head, you can set
% an abbreviated paper title here
%
\author{Dewen Zhang\inst{1}\orcidID{0009-0002-5145-0100} 
\and 
Tahir Hussain\inst{1}\orcidID{0009-0005-7937-6485} 
\and 
Wangpeng An\inst{2}\orcidID{0000-0002-1869-1837} 
\and 
Hayaru Shouno\inst{1}\orcidID{0000-0002-2412-0184}}
%
% \authorrunning{D. Zhang et al.}
% First names are abbreviated in the running head.
% If there are more than two authors, 'et al.' is used.
%
\institute{Department of Informatics, Graduate School of Informatics and Engineering, The University of Electro-Communications, Tokyo 182-8585, Japan\\
\email{\{zhangdewen,shouno\}@uec.ac.jp}\\
\email{(Tahir Hussain)f2240014@gl.cc.uec.ac.jp}\\
\and
TikTok Inc, 1199 Coleman Ave, San Jose, CA 95110\\
\email{anwangpeng@gmail.com}}
\maketitle              % typeset the header of the contribution
\begin{abstract}
Human pose estimation traditionally relies on architectures that encode keypoint priors, limiting their generalization to novel poses or unseen keypoints. Recent language-guided approaches like LocLLM reformulate keypoint localization as a vision-language task, enabling zero-shot generalization through textual descriptions. However, LocLLM's linear projector fails to capture complex spatial-textual interactions critical for high-precision localization. To address this, we propose PoseLLM, the first Large Language Model (LLM)-based pose estimation framework that replaces the linear projector with a nonlinear MLP vision-language connector. This lightweight two-layer MLP with GELU activation enables hierarchical cross-modal feature transformation, enhancing the fusion of visual patches and textual keypoint descriptions. Trained exclusively on COCO data, PoseLLM achieves 77.8 AP on the COCO validation set, outperforming LocLLM by +0.4 AP, while maintaining strong zero-shot generalization on Human-Art and MPII. Our work demonstrates that a simple yet powerful nonlinear connector significantly boosts localization accuracy without sacrificing generalization, advancing the state-of-the-art in language-guided pose estimation. Code is available at \url{https://github.com/Ody-trek/PoseLLM}.

\end{abstract}
\section{Introduction}
Human pose estimation, the task of localizing anatomical keypoints in images, underpins critical applications in augmented reality, human-computer interaction, sports analytics, and healthcare monitoring~\cite{27,28,29,30}. Traditional approaches predominantly fall into two categories: heatmap-based methods (e.g., HRNet~\cite{3}, ViTPose~\cite{4}) that predict spatial probability distributions, and regression-based methods (e.g., RLE~\cite{7}) that directly output coordinates. While effective, these methods inherently encode keypoint priors and spatial relationships within their network architectures or training data. This design fundamentally limits their ability to generalize to novel poses, unseen datasets, or new keypoint types not encountered during training.

Recently, the emergence of Vision-Language Models (VLMs) has opened new avenues for cross-modal reasoning. Pioneering work, LocLLM~\cite{5}, redefined keypoint localization by framing it as a vision-language question-answering task. By providing textual descriptions of keypoints (e.g., anatomical context and spatial relationships) alongside the image, LocLLM leverages the powerful reasoning capabilities of LLMs to achieve remarkable zero-shot generalization across datasets and keypoint types. This paradigm shift moves away from baking priors into the model architecture and instead utilizes explicit language guidance.

However, LocLLM relies on a simple linear projector to align visual features from the vision encoder with the input space of the Large Language Model (LLM). While efficient, this linear mapping constrains the model's capacity to capture the intricate, hierarchical, and often nonlinear interactions between localized visual features and the rich semantic and spatial cues present in the textual keypoint descriptions. This limitation potentially hinders the model's ability to achieve the highest precision in coordinate regression, a task demanding fine-grained spatial understanding.

To address this gap, we introduce PoseLLM, the first LLM-based pose estimation model that replaces the linear vision-language connector with a nonlinear Multi-Layer Perceptron (MLP). Our core insight is that a lightweight, learnable nonlinear transformation can significantly enhance the model's ability to fuse visual and textual information for precise localization. Specifically, PoseLLM employs a two-layer MLP with GELU activation~\cite{12} as the connector between the vision encoder and the LLM. This simple yet powerful modification enables richer hierarchical feature transformations, allowing the model to better model complex spatial dependencies between image regions and the textual descriptions of target keypoints.

We rigorously evaluate PoseLLM on standard benchmarks (COCO Keypoint~\cite{16}, Human-Art~\cite{17}, MPII~\cite{18}), training only on COCO data. PoseLLM consistently outperforms LocLLM, achieving a +0.4 AP improvement on the COCO validation set, demonstrating the efficacy of the nonlinear connector for boosting localization accuracy. Crucially, PoseLLM maintains the exceptional generalization capabilities characteristic of the language-guided paradigm, performing comparably or slightly better than LocLLM on cross-dataset evaluations (Human-Art, MPII) without any fine-tuning. This confirms that enhanced representational power through nonlinear alignment does not come at the cost of flexibility.

Our key contributions are:

\begin{itemize} [label=\textbullet]
\item We propose PoseLLM, the first LLM-based pose estimation model utilizing a nonlinear MLP-based vision-language connector, replacing the linear projector used in prior work like LocLLM.
\item We demonstrate that this simple architectural enhancement significantly improves localization precision (+0.4 AP on COCO) while preserving strong zero-shot cross-dataset generalization capabilities.
\item Our experiments establish new state-of-the-art results for language-guided pose estimation, providing a stronger baseline for future work in leveraging VLMs/LLMs for fine-grained spatial tasks.
\end{itemize}

\section{Related Work}
\subsection{Vision-Language Multimodal Models}
Recent advancements in vision-language multimodal models (VLMs) have demonstrated remarkable capabilities in cross-modal understanding and reasoning. Models such as $V^*$~\cite{23}, LLaVA~\cite{20,21}, BLIP-2~\cite{24}, Flamingo~\cite{25}, and DeepSeek-VL2~\cite{22} align visual inputs with LLMs to tackle diverse vision-language tasks. However, these models often struggle with fine-grained human-centric tasks such as pose estimation and action understanding due to insufficient spatial and structural priors in standard training data. To address this, recent works by Zhang et al.~\cite{10,11} propose keypoint-integrated instruction-following data generation, enriching traditional captions and bounding boxes with human keypoint annotations. This approach significantly improves pose-aware reasoning in VLMs but focuses primarily on descriptive and conversational tasks rather than quantitative keypoint localization.

\subsection{Human Pose Estimation}
Traditional human pose estimation methods fall into two categories: heatmap-based approaches (e.g., SimplePose~\cite{1}, SimCC~\cite{2}, HRNet~\cite{3}, ViTPose~\cite{4}), which predict probability maps for keypoints, and regression-based methods (e.g., DeepPose~\cite{6}, RLE~\cite{7}), which directly output coordinates. While effective, these models encode keypoint priors into their architecture, limiting generalization to unseen poses or novel keypoints. LocLLM~\cite{5} pioneered the use of LLMs for keypoint localization by reformulating it as a vision-language question-answering task. It utilizes textual descriptions of keypoints (e.g., anatomical context and spatial relationships) alongside images, enabling zero-shot generalization to new datasets and keypoint types. Despite its innovation, LocLLM employs a linear projector to align visual and textual features, which constrains its capacity to model complex cross-modal interactions critical for precise localization. In contrast, replacing this linear projector with an MLP enables richer hierarchical feature transformations, capturing intricate spatial relationships between visual patches and textual keypoint descriptions. This approach thereby enhances localization accuracy while maintaining generalization capabilities.

\subsection{MLP in Vision-Language Alignment}
MLP have proven effective as lightweight, nonlinear connectors in VLMs. In LLaVA-1.5~\cite{21}, an MLP projects image features into the LLM’s token space, enabling robust alignment for high-level reasoning tasks. Similarly, EarthDial~\cite{26}, a specialized VLM for remote sensing, employs an MLP connection block to align features from a lightweight visual encoder with a language model. Its design enables unified processing of multi-resolution, multi-spectral, and multi-temporal remote sensing data. However, these works focus on semantic alignment rather than spatial coordinate regression. To our knowledge, no prior work has explored nonlinear MLP connectors for keypoint coordinate prediction in an LLM-based framework. Building on LocLLM’s paradigm, we introduce PoseLLM, the first LLM-based model to replace the linear projector with a nonlinear MLP-based vision-language connector. This simple yet effective design captures intricate relationships between visual patches and keypoint descriptions, significantly enhancing localization precision. Our experiments establish stronger baselines, outperforming LocLLM.

\section{Method}
\begin{figure}[htbp]
    \centering
    \includegraphics[width=\textwidth]{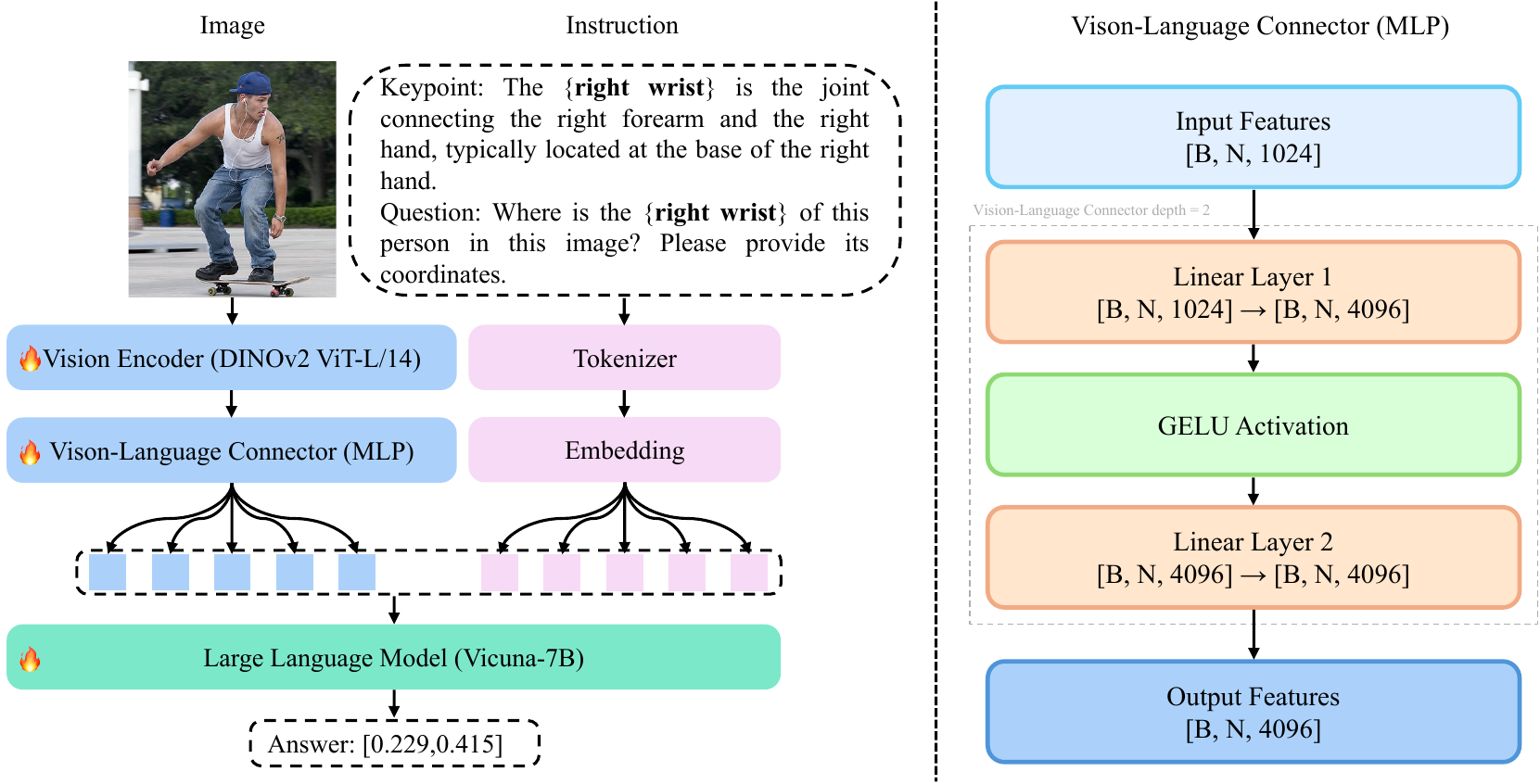} % 文件名改为您的PDF文件名
    \caption{Architecture of PoseLLM.}
    \label{figure 1}
\end{figure}

PoseLLM is designed to enhance the ability of vision-language models to perform precise keypoint localization and reasoning by introducing a nonlinear vision-language connector. The overall architecture of PoseLLM is illustrated in Fig.~\ref{figure 1}. It consists of three major components: a vision encoder, a vision-language connector (our contribution), and a LLM. The input to the model is an image along with an instruction describing a specific keypoint query. The model's goal is to infer the coordinates of the queried keypoint in the image.

\subsection{Vision Encoder}

We use the DINOv2 ViT-L/14~\cite{8} as the image encoder. Given an input image, the encoder outputs a sequence of patch-level image features:
\begin{equation}
    \mathbf{I} \in \mathbb{R}^{B \times N \times 1024},
\end{equation}
where $B$ is the batch size, $N$ is the number of visual tokens, and 1024 is the feature dimension of each token.

\subsection{Text Encoding}
The textual instruction provided to the model consists of two parts: a keypoint description and a corresponding question. The keypoint description offers anatomical and spatial context about a specific body part, while the question prompts the model to infer its location in the image. An example of such an instruction is shown in Fig.~\ref{figure 1}.

This instruction format is consistent with the original LocLLM design~\cite{5}, where such descriptive prompts are used to guide the LLM toward understanding the semantics and spatial properties of human body parts.

We tokenize the combined instruction using the Vicuna-7B tokenizer and then apply a standard embedding lookup to obtain a sequence of text embeddings:
\begin{equation}
    \mathbf{T} \in \mathbb{R}^{B \times L \times D},
\end{equation}
where $L$ is the number of tokens in the instruction prompt, and $D$ is the text embedding dimension compatible with the language model.

The detailed descriptions for all 17 keypoints used in our experiments are listed in Table~\ref{table 1}. These descriptions are identical to those used in LocLLM~\cite{5}, ensuring a fair comparison and consistent supervision.

\begin{table*}[htbp]
\caption{The list of keypoint location description.}
\centering
\begin{tcolorbox}[
    colback=gray!5,
    colframe=gray,
    boxrule=0.3mm,
    % arc=5mm,
    boxsep=0.8mm,
    left=2mm, right=2mm, top=1mm, bottom=1mm,
    fontupper=\footnotesize,
    rounded corners % sharp corners
]
\begin{itemize} [label=\textbullet]
\item  nose: The nose is the central, protruding feature on their face, located just above the upper lip.
\item  left eye: The left eye is the visual organ on the left side of their face, typically located above the left cheek and beside the nose.
\item  right eye: The right eye is the visual organ on the right side of their face, typically located above the right cheek and beside the nose.
\item  left ear: The left ear is the auditory organ on the left side of their head, typically located to the side of the left temple.
\item  right ear: The right ear is the auditory organ on the right side of their head, typically located to the side of the right temple.
\item  left shoulder: The left shoulder is the joint connecting the left arm and the torso, typically situated on the upper left side of the chest.
\item  right shoulder: The right shoulder is the joint connecting the right arm and the torso, typically situated on the upper right side of the chest.
\item  left elbow: The left elbow is the joint connecting the left upper arm and the left forearm, typically situated in the middle of the left arm, between left shoulder and left wrist.
\item  right elbow: The right elbow is the joint connecting the right upper arm and the right forearm, typically situated in the middle of the right arm, between right shoulder and right wrist.
\item  left wrist: The left wrist is the joint connecting the left forearm and the left hand, typically located at the base of the left hand.
\item  right wrist: The right wrist is the joint connecting the right forearm and the right hand, typically located at the base of the right hand.
\item  left hip: The left hip is the joint connecting the left thigh to the pelvis, typically located on the left side of the lower torso.
\item  right hip: The right hip is the joint connecting the right thigh to the pelvis, typically located on the right side of the lower torso.
\item  left knee: The left knee is the joint connecting the left thigh and the left lower leg, typically situated in the middle of the left leg, it is located between the left hip and left ankle.
\item  right knee: The right knee is the joint connecting the upper leg and lower leg on the right side, it is located between the right hip and right ankle.
\item  left ankle: The left ankle is the joint connecting the left lower leg and the left foot, typically located at the base of the left leg.
\item  right ankle: The right ankle is the joint connecting the right lower leg and the right foot, typically located at the base of the right leg.
\end{itemize}
\end{tcolorbox}
\label{table 1}
\end{table*}

\subsection{Nonlinear Vision-Language Connector (MLP)}

To bridge the modality gap between the visual embeddings and the LLM, we introduce a \textbf{nonlinear connector module} — a two-layer MLP with intermediate GELU activation~\cite{12}. Unlike LocLLM that relies on a single linear projection, our design enables richer cross-modal transformations through nonlinearity.

Given visual features $\mathbf{I} \in \mathbb{R}^{B \times N \times 1024}$, the connector transforms them as follows:
\begin{align}
    \mathbf{Z} &= \text{GELU}(\mathbf{I} \cdot \mathbf{W}_1 + \mathbf{b}_1), \tag{3} \\
    \mathbf{V} &= \mathbf{Z} \cdot \mathbf{W}_2 + \mathbf{b}_2, \tag{4}
\end{align}
where $\mathbf{W}_1 \in \mathbb{R}^{1024 \times 4096}$, $\mathbf{b}_1 \in \mathbb{R}^{4096}$, $\mathbf{W}_2 \in \mathbb{R}^{4096 \times 4096}$, and $\mathbf{b}_2 \in \mathbb{R}^{4096}$ are learnable parameters in the MLP. The resulting output $\mathbf{V} \in \mathbb{R}^{B \times N \times 4096}$ is the transformed visual feature sequence. This output is then concatenated with the text embeddings $\mathbf{T}$ to form the multimodal input to the LLM.

We show in experiments that this seemingly simple enhancement leads to better generalization and reasoning on human pose understanding tasks.

\subsection{Large Language Model and Output}

We adopt Vicuna-7B~\cite{9} as our backbone language model. It takes the concatenated multimodal sequence $[\mathbf{V}; \mathbf{T}]$ as input and performs autoregressive decoding to generate coordinate outputs corresponding to the queried keypoint.

To align the format of output coordinates with supervised ground truth, we train PoseLLM using teacher-forcing with ground-truth coordinates as targets. The model is optimized using the cross-entropy loss, computed only on the predicted answer tokens.

\section{Experiments}

To validate the effectiveness of our proposed nonlinear vision-language connector, we conduct comprehensive experiments on keypoint localization and cross-dataset generalization tasks. Following the experimental settings in LocLLM~\cite{5}, we adopt COCO~\cite{16}, Human-Art~\cite{17}, and MPII~\cite{18} datasets for evaluation. Our model is trained only on the COCO Keypoint training set, without the use of additional datasets, ensuring a fair comparison with previous state-of-the-art methods.

Specifically, we use the COCO Keypoint dataset for training, which contains approximately 57K images and 150K annotated human instances labeled with 17 keypoints. The Human-Art and MPII datasets are utilized exclusively for evaluating our model’s cross-dataset generalization capabilities.

Evaluation metrics follow the standard practices in human pose estimation tasks. On the COCO Keypoint dataset, we adopt the mean Average Precision (mAP), along with AP50, AP75, APM, and APL metrics, reflecting precision under various object scales and IoU thresholds. For the MPII dataset, we report PCKh@0.5 and PCKh@0.1, which measure localization accuracy at different thresholds relative to head size. The Human-Art dataset evaluation follows similar protocols to COCO, employing the standard mAP and associated precision metrics.

Our PoseLLM model introduces a nonlinear vision-language connector comprising a two-layer MLP with a GELU activation function, replacing the single linear projection in the original LocLLM. We train our model for 12 epochs using two NVIDIA A6000 GPUs, employing the AdamW optimizer with a learning rate of 5e-4, a batch size of 32 via gradient accumulation, and a weight decay of 0.05. The input images are uniformly resized to 224×224 pixels. We fine-tune the vision encoder (DINOv2 ViT-L/14) and the LLM (Vicuna-7B) using LoRA modules~\cite{19}, while fully updating the parameters of the vision-language connector (MLP). This parameter-efficient strategy ensures effective learning while maintaining computational feasibility.

\subsection{Qualitative Evaluation}

\begin{figure}[htbp]
    \centering
    \includegraphics[width=\textwidth]{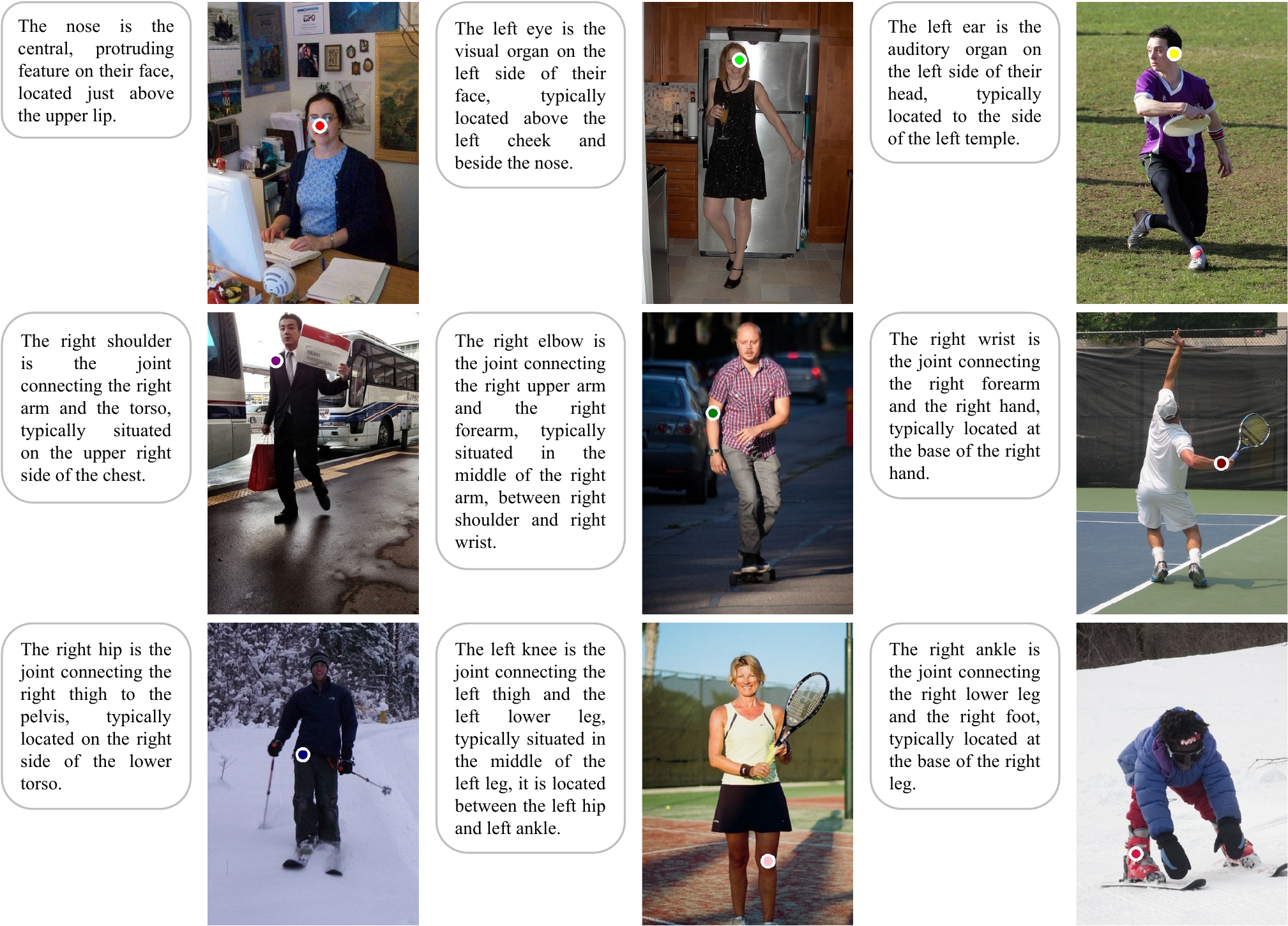} % 文件名改为您的PDF文件名
    \caption{Qualitative results of PoseLLM. Each example displays the input image with the predicted keypoint location and its associated descriptive prompt.}
    \label{figure 2}
\end{figure}

Fig.~\ref{figure 2} illustrates several examples from the COCO validation set, showing our model's predictions across various keypoints. Each image is accompanied by a descriptive instruction and a marked keypoint location predicted by PoseLLM. The examples demonstrate the model’s ability to comprehend spatial and anatomical information encoded in the instruction and accurately infer the corresponding keypoint location.

\subsection{Quantitative Evaluation}

\paragraph{COCO Keypoint Localization.} Table~\ref{table 2} compares our model with existing heatmap-based, regression-based, and language-based methods on the COCO validation set. PoseLLM achieves competitive performance with state-of-the-art methods, outperforming baseline LocLLM by +0.4 in AP, demonstrating the benefit of our nonlinear vision-language connector.

\begin{table}[htbp]
\centering
\caption{Comparison with other methods on COCO Keypoint validation set. All results are obtained using the official model weights and evaluated with ground-truth bounding boxes, without flip test.}
% \small
\setlength{\tabcolsep}{5pt}
\begin{tabular}{lcccccc}
\toprule
Method & AP & AP$^{50}$ & AP$^{75}$ & AP$^{M}$ & AP$^{L}$ & AR \\
\midrule
\multicolumn{7}{l}{\textit{Heatmap-based}} \\
\midrule
SimplePose~\cite{1} & 74.4 & 92.6 & 82.5 & 71.5 & 79.2 & 77.6 \\
SimCC~\cite{2}       & 76.5 & 93.2 & 83.1 & 73.6 & 81.5 & 79.7 \\
HRNet~\cite{3}       & 76.8 & 93.6 & 83.6 & 74.0 & 81.5 & 79.6 \\
ViTPose~\cite{4}   & 77.4 & 93.6 & 84.8 & 74.7 & 81.9 & 80.2 \\
\midrule
\multicolumn{7}{l}{\textit{Regression-based}} \\
\midrule
DeepPose~\cite{6} & 53.8 & 82.6 & 59.2 & 52.2 & 57.3 & 66.8 \\
RLE~\cite{7}      & 74.0 & 91.5 & 81.6 & 70.9 & 78.5 & 76.8 \\
\midrule
\multicolumn{7}{l}{\textit{Language-based}} \\
\midrule
CLIP baseline~\cite{5}    & 73.1 & 92.5 & 81.3 & 70.3 & 77.4 & 76.5 \\
LocLLM~\cite{5}  & 77.4 & 94.4 & 85.2 & 74.5 & 81.8 & 80.6 \\
\textbf{PoseLLM (Ours)}  & \textbf{77.8} & \textbf{94.4} & \textbf{85.4} & \textbf{74.7} & \textbf{82.9} & \textbf{80.8} \\
\bottomrule
\end{tabular}
\label{table 2}
\end{table}

\paragraph{Cross-Dataset Generalization.} To evaluate the generalization ability of our model, we test PoseLLM trained on COCO directly on Human-Art and MPII datasets without any finetuning. Table~\ref{table 3} shows that PoseLLM achieves comparable results with LocLLM on both datasets, with slight improvements in several keypoints. This indicates that our nonlinear connector does not compromise generalization while enhancing representation learning.

\begin{table}[htbp]
\centering
\caption{Cross-dataset generalization performance of PoseLLM on Human-Art and MPII. All models are trained on the COCO Keypoint training set and evaluated without finetuning. We report accuracy only for keypoints that appear in COCO Keypoint.}
% \small
\setlength{\tabcolsep}{4pt} % 调整列间距，避免过宽或过窄
\begin{tabular}{l|cccccc|cccccc}
% \begin{tabularx}{\textwidth}{l|*{6}{>{\centering\arraybackslash}X}|*{6}{>{\centering\arraybackslash}X}}

\toprule
\multirow{2}{*}{\raisebox{-0.6ex}{\centering Method}} & \multicolumn{6}{c|}{Human-Art} & \multicolumn{6}{c}{MPII} \\
\cmidrule{2-13}
& AP & AP$^{50}$ & AP$^{75}$ & AP$^{M}$ & AP$^{L}$ & AR & Shou. & Elbo. & Hip & Knee & Mean & Mean0.1 \\
\midrule
SimplePose \cite{1} & 48.4 & 73.0 & 50.7 & 27.2 & 50.7 & 52.8 & 93.7 & 85.6 & 85.4 & 81.6 & 84.7 & 22.2 \\
SimCC \cite{2} & 51.7 & 75.2 & 54.8 & 26.2 & 54.3 & 57.0 & 92.2 & 84.1 & 82.8 & 80.5 & 83.2 & 28.5 \\
HRNet \cite{3} & 53.4 & 76.3 & 56.5 & 30.4 & 55.9 & 57.5 & 93.4 & 86.1 & 85.0 & 81.9 & 85.8 & 26.9 \\
ViTPose \cite{4} & 53.8 & 77.9 & 57.4 & 31.4 & 56.6 & 58.7 & 94.5 & 88.2 & 87.3 & 85.0 & 86.9 & 25.8 \\
CLIP baseline \cite{5} & 49.3 & 75.7 & 51.8 & 27.7 & 52.0 & 54.5 & 95.5 & 88.8 & 87.5 & 84.7 & 87.1 & 25.0 \\
LocLLM \cite{5} & 67.8 & \textbf{88.6} & 73.9 & 41.6 & 70.8 & 71.9 & 98.9 & \textbf{98.2} & 99.2 & 98.4 & 98.1 & 80.6 \\
\textbf{PoseLLM (Ours)} & \textbf{67.9} & 87.7 & \textbf{74.2} & \textbf{41.7} & \textbf{71.1} & \textbf{72.0} & \textbf{98.9} & 98.1 & \textbf{99.3} & \textbf{98.4} & \textbf{98.1} & \textbf{80.7} \\
\bottomrule
\end{tabular}
\label{table 3}
\end{table}

\section{Conclusion}
We present PoseLLM, a novel paradigm for human pose estimation that integrates a nonlinear MLP connector into an LLM-based vision-language framework. By replacing LocLLM’s linear projector with a two-layer MLP, PoseLLM captures intricate spatial relationships between image patches and textual keypoint descriptions, enabling richer cross-modal reasoning. Extensive experiments validate our design: PoseLLM achieves state-of-the-art performance (77.8 AP) on the COCO benchmark, surpassing LocLLM by +0.4 AP. Crucially, PoseLLM retains LocLLM’s zero-shot generalization capabilities, matching or slightly improving performance on Human-Art and MPII without fine-tuning. This confirms that nonlinear alignment enhances precision while preserving flexibility.

Our work highlights the untapped potential of lightweight nonlinear connectors in vision-language models for fine-grained spatial tasks. Future research could explore dynamic MLP designs, and integration with larger multimodal LLMs.

\subsubsection{Acknowledgements} This work was supported by JST SPRING, Grant Number JPMJSP2131.

%
% ---- Bibliography ----
%
% BibTeX users should specify bibliography style 'splncs04'.
% References will then be sorted and formatted in the correct style.
%
\bibliographystyle{splncs04}
\bibliography{mybibliography}

\begin{thebibliography}{10}
\providecommand{\url}[1]{\texttt{#1}}
\providecommand{\urlprefix}{URL }
\providecommand{\doi}[1]{https://doi.org/#1}

\bibitem{25}
Alayrac, J.B., Donahue, J., Luc, P., Miech, A., Barr, I., Hasson, Y., Lenc, K., Mensch, A., Millican, K., Reynolds, M., et~al.: Flamingo: a visual language model for few-shot learning. Advances in neural information processing systems  \textbf{35},  23716--23736 (2022)

\bibitem{18}
Andriluka, M., Pishchulin, L., Gehler, P., Schiele, B.: 2d human pose estimation: New benchmark and state of the art analysis. In: Proceedings of the IEEE Conference on computer Vision and Pattern Recognition. pp. 3686--3693 (2014)

\bibitem{28}
Avogaro, A., Cunico, F., Rosenhahn, B., Setti, F.: Markerless human pose estimation for biomedical applications: a survey. Frontiers in Computer Science  \textbf{5},  1153160 (2023)

\bibitem{30}
Chen, J., Hu, J., Wang, G., Jiang, Z., Zhou, T., Chen, Z., Lv, C.: Taoavatar: Real-time lifelike full-body talking avatars for augmented reality via 3d gaussian splatting. In: Proceedings of the Computer Vision and Pattern Recognition Conference. pp. 10723--10734 (2025)

\bibitem{9}
Chiang, W.L., Li, Z., Lin, Z., Sheng, Y., Wu, Z., Zhang, H., Zheng, L., Zhuang, S., Zhuang, Y., Gonzalez, J.E., Stoica, I., Xing, E.P.: Vicuna: An open-source chatbot impressing gpt-4 with 90\%* chatgpt quality (March 2023), \url{https://lmsys.org/blog/2023-03-30-vicuna/}

\bibitem{12}
Hendrycks, D., Gimpel, K.: Gaussian error linear units (gelus). arXiv preprint arXiv:1606.08415  (2016)

\bibitem{19}
Hu, E.J., Shen, Y., Wallis, P., Allen-Zhu, Z., Li, Y., Wang, S., Wang, L., Chen, W.: Lora: Low-rank adaptation of large language models. ICLR  \textbf{1}(2), ~3 (2022)

\bibitem{17}
Ju, X., Zeng, A., Wang, J., Xu, Q., Zhang, L.: Human-art: A versatile human-centric dataset bridging natural and artificial scenes. In: Proceedings of the IEEE/CVF conference on computer vision and pattern recognition. pp. 618--629 (2023)

\bibitem{7}
Li, J., Bian, S., Zeng, A., Wang, C., Pang, B., Liu, W., Lu, C.: Human pose regression with residual log-likelihood estimation. In: Proceedings of the IEEE/CVF international conference on computer vision. pp. 11025--11034 (2021)

\bibitem{24}
Li, J., Li, D., Savarese, S., Hoi, S.: Blip-2: Bootstrapping language-image pre-training with frozen image encoders and large language models. In: International conference on machine learning. pp. 19730--19742. PMLR (2023)

\bibitem{2}
Li, Y., Yang, S., Liu, P., Zhang, S., Wang, Y., Wang, Z., Yang, W., Xia, S.T.: Simcc: A simple coordinate classification perspective for human pose estimation. In: European conference on computer vision. pp. 89--106. Springer (2022)

\bibitem{16}
Lin, T.Y., Maire, M., Belongie, S., Hays, J., Perona, P., Ramanan, D., Doll{\'a}r, P., Zitnick, C.L.: Microsoft coco: Common objects in context. In: Computer vision--ECCV 2014: 13th European conference, zurich, Switzerland, September 6-12, 2014, proceedings, part v 13. pp. 740--755. Springer (2014)

\bibitem{21}
Liu, H., Li, C., Li, Y., Lee, Y.J.: Improved baselines with visual instruction tuning. In: Proceedings of the IEEE/CVF Conference on Computer Vision and Pattern Recognition. pp. 26296--26306 (2024)

\bibitem{20}
Liu, H., Li, C., Wu, Q., Lee, Y.J.: Visual instruction tuning. Advances in neural information processing systems  \textbf{36},  34892--34916 (2023)

\bibitem{8}
Oquab, M., Darcet, T., Moutakanni, T., Vo, H., Szafraniec, M., Khalidov, V., Fernandez, P., Haziza, D., Massa, F., El-Nouby, A., et~al.: Dinov2: Learning robust visual features without supervision. arXiv preprint arXiv:2304.07193  (2023)

\bibitem{26}
Soni, S., Dudhane, A., Debary, H., Fiaz, M., Munir, M.A., Danish, M.S., Fraccaro, P., Watson, C.D., Klein, L.J., Khan, F.S., et~al.: Earthdial: Turning multi-sensory earth observations to interactive dialogues. In: Proceedings of the Computer Vision and Pattern Recognition Conference. pp. 14303--14313 (2025)

\bibitem{3}
Sun, K., Xiao, B., Liu, D., Wang, J.: Deep high-resolution representation learning for human pose estimation. In: Proceedings of the IEEE/CVF conference on computer vision and pattern recognition. pp. 5693--5703 (2019)

\bibitem{6}
Toshev, A., Szegedy, C.: Deeppose: Human pose estimation via deep neural networks. In: Proceedings of the IEEE conference on computer vision and pattern recognition. pp. 1653--1660 (2014)

\bibitem{5}
Wang, D., Xuan, S., Zhang, S.: Locllm: Exploiting generalizable human keypoint localization via large language model. In: Proceedings of the IEEE/CVF Conference on Computer Vision and Pattern Recognition (CVPR) (2024)

\bibitem{23}
Wu, P., Xie, S.: V?: Guided visual search as a core mechanism in multimodal llms. In: Proceedings of the IEEE/CVF Conference on Computer Vision and Pattern Recognition. pp. 13084--13094 (2024)

\bibitem{22}
Wu, Z., Chen, X., Pan, Z., Liu, X., Liu, W., Dai, D., Gao, H., Ma, Y., Wu, C., Wang, B., et~al.: Deepseek-vl2: Mixture-of-experts vision-language models for advanced multimodal understanding. arXiv preprint arXiv:2412.10302  (2024)

\bibitem{29}
Xi, X., Zhang, C., Jia, W., Jiang, R.: Enhancing human pose estimation in sports training: Integrating spatiotemporal transformer for improved accuracy and real-time performance. Alexandria Engineering Journal  \textbf{109},  144--156 (2024)

\bibitem{1}
Xiao, B., Wu, H., Wei, Y.: Simple baselines for human pose estimation and tracking. In: Proceedings of the European conference on computer vision (ECCV). pp. 466--481 (2018)

\bibitem{4}
Xu, Y., Zhang, J., Zhang, Q., Tao, D.: Vitpose: Simple vision transformer baselines for human pose estimation. Advances in neural information processing systems  \textbf{35},  38571--38584 (2022)

\bibitem{10}
Zhang, D., An, W., Shouno, H.: Keypoint-integrated instruction-following data generation for enhanced human pose and action understanding in multimodal models (2025), \url{https://arxiv.org/abs/2409.09306}

\bibitem{11}
Zhang, D., Hussain, T., An, W., Shouno, H.: Llava-pose: Enhancing human pose and action understanding via keypoint-integrated instruction tuning (2025), \url{https://arxiv.org/abs/2506.21317}

\bibitem{27}
Zheng, C., Wu, W., Chen, C., Yang, T., Zhu, S., Shen, J., Kehtarnavaz, N., Shah, M.: Deep learning-based human pose estimation: A survey. ACM Computing Surveys  \textbf{56}(1),  1--37 (2023)

\end{thebibliography}

\end{document}